\documentclass[11pt,a4paper]{article}
\usepackage[hyperref]{naaclhlt2018}
\usepackage{times}
\usepackage{latexsym}
\usepackage{graphicx}
\usepackage{url}
\usepackage[utf8]{inputenc}
\usepackage{color}
\usepackage{xcolor}
\usepackage{booktabs}
\usepackage{tipa}
\usepackage{fancyvrb}
\usepackage{multirow}
\definecolor{nice-red}{HTML}{E41A1C}
\definecolor{nice-orange}{HTML}{FF7F00}
\definecolor{nice-yellow}{HTML}{FFC020}
\definecolor{nice-green}{HTML}{4DAF4A}
\definecolor{nice-blue}{HTML}{377EB8}
\definecolor{nice-purple}{HTML}{984EA3}

\aclfinalcopy 
\def\aclpaperid{245} 

\title{From Phonology to Syntax: Unsupervised Linguistic Typology\\ at Different Levels with Language Embeddings}

\author{Johannes Bjerva\\
Department of Computer Science\\
University of Copenhagen\\
Denmark\\
{\tt bjerva@di.ku.dk} \\\And
Isabelle Augenstein\\
Department of Computer Science\\
University of Copenhagen\\
Denmark\\
{\tt augenstein@di.ku.dk}
\\}

\date{}

\begin{document}
\maketitle
\begin{abstract}
A core part of linguistic typology is the classification of languages according to linguistic properties, such as those detailed in the World Atlas of Language Structure (WALS).
Doing this manually is prohibitively time-consuming, which is in part evidenced by the fact that only 100 out of over 7,000 languages spoken in the world are fully covered in WALS.
%

We learn distributed language representations, 
which can be used to predict typological properties on a massively multilingual scale. 
Additionally, quantitative and qualitative analyses of these language embeddings can tell us how language similarities are encoded in NLP models for tasks at different typological levels.
The representations are learned in an unsupervised manner alongside tasks at three typological levels: phonology (grapheme-to-phoneme prediction, and phoneme reconstruction), morphology (morphological inflection), and syntax (part-of-speech tagging). 

We consider more than 800 languages and find significant differences in the language representations encoded, depending on the target task.
For instance, although Norwegian Bokmål and Danish are typologically close to one another, they are phonologically distant, which is reflected in their language embeddings growing relatively distant in a phonological task. 
We are also able to predict typological features in WALS with high accuracies, even for unseen language families.
\end{abstract}

\section{Introduction}
For more than two and a half centuries, linguistic typologists have studied languages with respect to their structural and functional properties 
\citep{haspelmath:2001,velupillai:2012}.
Although typology has a long history \citep{herder:1772,gabelentz:1891,greenberg:1960,greenberg:1974,dahl:1985,comrie:1989,haspelmath:2001,croft:2002}, computational approaches have only recently gained popularity \citep{dunn:2011,walchli:2014,ostling:2015,
cotterell:2017,asgari:typology,malaviya:2017,bjerva:2017:uralic}. 
One part of traditional typological research can be seen as assigning sparse explicit feature vectors to languages, for instance manually encoded in databases such as the World Atlas of Language Structures (WALS, \citealp{wals}).
A recent development which can be seen as analogous to this is the process of learning distributed language representations in the form of dense real-valued vectors, often referred to as \textit{language embeddings} \citep{tsvetkov:2016,ostling_tiedemann:2017,malaviya:2017}.
We hypothesise that these language embeddings encode typological properties of language, reminiscent of the features in WALS, or even of parameters in Chomsky's Principles and Parameters framework \citep{chomsky:1993}.

In this paper, we model languages in deep neural networks using language embeddings, considering three typological levels: phonology, morphology and syntax. 
We consider four NLP tasks to be representative of these levels: grapheme-to-phoneme prediction and phoneme reconstruction, morphological inflection, and part-of-speech tagging.
We pose three research questions (\textbf{RQ}s):
\begin{enumerate}
\setlength{\itemindent}{.29cm}
    \item [{\bf RQ 1}] Which typological properties are encoded in task-specific distributed language representations, and can we predict phonological, morphological and syntactical properties of languages using such representations?
    \item [{\bf RQ 2}] To what extent do the encoded properties change as the representations are fine tuned for tasks at different linguistic levels?
    \item [{\bf RQ 3}] How are language similarities encoded in fine-tuned language embeddings?
\end{enumerate}
One of our key findings is that language representations differ considerably depending on the target task. 
For instance, for grapheme-to-phoneme mapping, the differences between the representations for Norwegian Bokmål and Danish increase rapidly during training.
This is due to the fact that, although the languages are typologically close to one another, they are phonologically distant.

\section{Related work}
Computational linguistics approaches to typology are now possible on a larger scale than ever before due to advances in neural computational models.
Even so, recent work only deals with fragments of typology compared to what we consider here.

\paragraph{Computational typology}
has to a large extent focused on exploiting word or morpheme alignments on the massively parallel New Testament, in approximately 1,000 languages, in order to infer word-order \citep{ostling:2015} or assign linguistic categories \citep{asgari:typology}.
\citet{walchli:2014} similarly extracts lexical and grammatical markers using New Testament data.
Other work has taken a computational perspective on language evolution \citep{dunn:2011}, and phonology \citep{cotterell:2017,alishahi:2017}.

\paragraph{Language embeddings}
In this paper, we follow an approach which has seen some attention in the past year, namely the use of distributed language representations, or \textit{language embeddings}.
Some typological experiments are carried out by \citet{ostling_tiedemann:2017}, who learn language embeddings via multilingual language modelling and show that they can be used to reconstruct genealogical trees.
\citet{malaviya:2017} learn language embeddings via neural machine translation, and predict word-ordering features.

\paragraph{Contributions} Our work bears the most resemblance to \citet{bjerva:2017:uralic}, who fine-tune language embeddings on the task of PoS tagging, and investigate how a handful of typological properties are coded in these for four Uralic languages.
We expand on this and thus contribute to previous work by:
(i) introducing novel qualitative investigations of language embeddings, in addition to thorough quantitative evaluations;
(ii) considering four tasks at three different typological levels;
(iii) considering a far larger sample of several hundred languages; and 
(iv) grounding the language representations in linguistic theory.

\section{Background}
\subsection{Distributed language representations}
There are several methods for obtaining distributed language representations by training a recurrent neural language model \citep{rnnlm} simultaneously for different languages \citep{tsvetkov:2016,ostling_tiedemann:2017}.
In these recurrent multilingual language models with long short-term memory cells (LSTM, \citealp{lstm}), languages are embedded into a $n$-dimensional space.
In order for multilingual parameter sharing to be successful in this setting, the neural network is encouraged to use the language embeddings to encode features of language.
In this paper, we explore the embeddings trained by \citet{ostling_tiedemann:2017}, both in their original state, and by further tuning them for our four tasks. 
While other work has looked at the types of representations encoded in different layers of deep neural models \citep{kadar:2017}, we choose to look at the representations only in the bottom-most embedding layer.
This is motivated by the fact that we look at several tasks using different neural architectures, and want to ensure comparability between these.

\subsubsection{Language embeddings as continuous Chomskyan parameter vectors}
We now turn to the theoretical motivation of the language representations.
The field of NLP is littered with distributional word representations, which are theoretically justified by distributional semantics \citep{Harris:1954,firth}, summarised in the catchy phrase \textit{You shall know a word by the company it keeps} \citep{firth}.
We argue that language embeddings, or distributed representations of language, can also be theoretically motivated by Chomsky's Principles and Parameters framework \citep{chomskylesnik:1993,chomsky:1993,chomsky:2014}.
Language embeddings encode languages as dense real-valued vectors, in which the dimensions are reminiscent of the parameters found in this framework. 
Briefly put, Chomsky argues that languages can be described in terms of principles (abstract rules) and parameters (switches) which can be turned either on or off for a given language \citep{chomskylesnik:1993}.
An example of such a switch might represent the positioning of the head of a clause (i.e. either head-initial or head-final). 
For English, this switch would be set to the `initial' state, whereas for Japanese it would be set to the `final' state. 
Each dimension in an $n$-dimensional language embedding might also describe such a switch, albeit in a more continuous fashion. 
The number of dimensions used in our language representations, $64$, is a plausible number of parameter vector dimensions \citep{dunn:2011}.
If we were able to predict typological features using such representations, this lends support to the argument that languages, at the very least, \textit{can} be represented by theoretically motivated parameter vectors, with the given dimensionality.

\subsection{Typological features in the World Atlas of Language Structure}
In the experiments for \textbf{RQ1} and \textbf{RQ2} we predict typological features extracted from WALS \citep{wals}.
We choose to investigate three linguistic levels of language: phonology, morphology, and syntax.
This is motivated by three factors:
(i) these features are related to NLP tasks for which data is available for a large language sample;
(ii) the levels cover a range from basic phonological and morphological structure, to syntactic structure, allowing us to approach our research question from several angles; and
(iii) the features in these categories are coded in WALS for a relatively large selection of languages.
We extract the three feature sets which represent these typological levels of language from WALS.\footnote{These are defined in the chapter structure in WALS: \url{http://wals.info/chapter}}

\paragraph{Phonological features}
cover $20$ features ranging from descriptions of the consonant and vowel inventories of a particular language to presence of tone and stress markers.
As an example, consider WALS feature 13A (Tone).\footnote{\url{http://wals.info/feature/13A}}
This feature takes three feature values: (i) \textit{no tones}, (ii) \textit{simple tone system}, and (iii) \textit{complex tone system}. 
Most Indo-European languages, such as English, Spanish, and Russian, do not have any tones (i).
Norwegian and Swedish are exceptions to this, as they both have simple tone systems (ii) similar to that in Japanese. 
Finally, complex tone systems (iii) are typically found in several African languages as well as languages in South-East Asia.

\paragraph{Morphological features}
cover a total of $41$ features.
We consider the features included in the Morphology chapter as well as those included in the Nominal Categories chapter to be morphological in nature.\footnote{This choice was made as, e.g., feature 37A (Definite Articles) includes as a feature value whether a definite affix is used.}
This includes features such as the number of genders, usage of definite and indefinite articles and reduplication.
As an example, consider WALS feature 37A (Definite Articles).\footnote{\url{http://wals.info/feature/37A}}
This feature takes five values:
(i) \textit{Definite word distinct from demonstrative}, 
(ii) \textit{Demonstrative word used as definite article}, 
(iii) \textit{Definite affix}, 
(iv) \textit{No definite, but indefinite article}, 
and (v) \textit{No definite or indefinite article}.
Again, most Indo-European languages fall into category (i), with Norwegian, Swedish, and Danish as relative outliers in category (iii).

\paragraph{Word-order features}
cover $56$ features, encoding properties such as the ordering of subjects, objects and verbs.
As an example, consider WALS feature 81A (Order of Subject, Object and Verb).\footnote{\url{http://wals.info/feature/81A}}
This feature takes all possible combinations of the three word classes as its feature values, with the addition of a special class for \textit{No dominant order}.
Most languages in WALS fall into the categories SOV (41.0\%) and SVO (35.4\%).


\section{Method}
The general set-up of the experiments in this paper are as follows.
We aim at answering our three research questions dealing with typological properties and similarities as encoded in language embeddings.
In order to do this, we attempt to predict typological features as they are encoded in WALS, using language embeddings which have been fine-tuned during training on tasks related to different typological properties.
The main interest in this paper is therefore not on how well each model performs on a given NLP task, but rather on what the resulting language embeddings encode.

Concretely, we use language embeddings $\vec{l}_i$ from a given training iteration of a given task as input to a k-NN classifier, which outputs the typological class a language belongs to (as coded in WALS).
We train separate classifiers for each typological property and each target task.
When $i=0$, this indicates the pre-trained language embeddings as obtained from \citet{ostling_tiedemann:2017}.
Increasing $i$ indicates the number of iterations over which the system at hand has been trained.
In each experiment, for a given iteration $i$, we consider each $\vec{l}_i\in L$ where $L$ is the intersection $L_{task}\cap L_{pre}$, where $L_{task}$ is the set of languages for a given task, and $L_{pre}$ is the set of the languages for which we have pre-trained embeddings.

All results in the following sections are the mean of three-fold cross-validation, and the mean over the WALS features in each given category.\footnote{The mean accuracy score is a harsh metric, as some features are very difficult to predict due to them, e.g., being very language specific or taking a large number of different values.}
We run the experiments in a total of three settings:
(i) evaluating on randomly selected language/feature pairs from a task-related feature set;
(ii) evaluating on an \textit{unseen language family from a task-related feature set};
(iii) evaluating on randomly selected language/feature pairs from all WALS feature sets.
This allows us to establish how well we can predict task-related features given a random sample of languages (i), and a sample from which a whole language family has been omitted (ii).
Finally, (iii) allows us to compare the task-specific feature encoding with a general one.

A baseline reference is also included, which is defined as the most frequently occurring typological trait within each category.\footnote{The languages represented in several of the tasks under consideration have a high Indo-European bias. Hence, several of the properties are relatively uniformly distributed, providing us with a strong baseline.}
For instance, in the morphological experiments, we only consider the $41$ WALS features associated with the categories of morphology and nominal categories.
The overlap between languages for which we have data for morphological inflection and languages for which these WALS features are coded is relatively small (fewer than $20$ languages per feature).
This small dataset size is why we have opted for a non-parametric $k$-Nearest Neighbours classifier for the typological experiments.
We use $k=1$, as several of the features take a large number of class values, and might only have a single instance represented in the training set.

Table~\ref{tab:overview} shows the datasets we consider (detailed in later sections), the typological class they are related to, the size of the language sample in the task, and the size of the intersection $L_{task}\cap L_{pre}$.
The number of pre-trained language embeddings, $|L_{pre}|$, is 975 in all cases.
We focus the evaluation for each task-specific language embedding set on the typological property relevant to that dataset.
In addition, we also evaluate on a set of all typological properties in WALS.
Note that the evaluation on all properties is only comparable to the evaluation on each specific property, as the set of languages under consideration differs between tasks.

\begin{table}[htbp]
  \resizebox{\columnwidth}{!}{
    \begin{tabular}{llrr}
    \toprule
    \textbf{Dataset} & \textbf{Class} & \textbf{$|L_{task}|$} & \textbf{$|L_{task}\cap L_{pre}|$} \\
    \midrule
    G2P        & Phonology     & 311   & 102 \\
    ASJP       & Phonology     & 4,664 & 824 \\
    SIGMORPHON & Morphology    & 52    & 29 \\
    UD         & Syntax        & 50    & 27 \\
    \bottomrule
    \end{tabular}
  }
  \caption{\label{tab:overview}Overview of tasks and datasets.}
\end{table}

\section{Phonology}
\subsection{Grapheme-to-phoneme}
We use grapheme-to-phoneme (G2P) as a proxy of a phonological task \citep{deri:2016,peters:2017}.
The dataset contains over 650,000 such training instances, for a total of 311 languages \citep{deri:2016}.
The task is to produce a phonological form of a word, given its orthographic form and the language in which it is written.
Crucially, this mapping is highly different depending on the language at hand.
For instance, take the word written \textit{variation}, which exists in both English and French:

\begin{Verbatim}[commandchars=\\\{\},fontsize=\small]
(English, variation) -> \textipa{""vE@ri"eIS@n}
(French,  variation) -> \textipa{""vaKja"sj\~O}
\end{Verbatim}

\subsubsection{Experiments and Analysis}
We train a sequence-to-sequence model with attention for the task of grapheme-to-phoneme mapping.\footnote{The system is described in detail in Section~\ref{sec:implementation}.}
The model takes as input the characters of each source form together with the language embedding for the language at hand and outputs a predicted phonological form.
Input and output alphabets are shared across all languages.
The system is trained over 3,000 iterations.

\paragraph{Quantitative results}
Since we consider Grapheme-to-Phoneme as a phonological task, we focus the quantitative evaluation on phonological features from WALS.
We run experiments using the language embeddings as features for a simple k-NN classifier.
The results in Table~\ref{tab:g2p} indicate that G2P is a poor proxy for language phonology, however, as typological properties pertaining to phonology are not encoded.
That is to say, the k-NN results do not outperform the baseline, and performance is on par even after fine tuning (no significant difference, $p>0.05$).
In the unseen setting, however, we find that pre-trained language embeddings are significantly better ($p>0.05$) at predicting the phonological features than both fine-tuned ones and the baseline.

\begin{table}[htbp]
  \centering
  \resizebox{\columnwidth}{!}{
    \begin{tabular}{lrrr}
      \toprule
      \textbf{System / features} & \textbf{Random phon.} & \textbf{Unseen phon.} & \textbf{All feat.} \\
      \midrule
      Most Frequent Class & *\textbf{75.46}\% & 65.57\% & 79.90\% \\
      k-NN (pre-trained)  & 71.45\%           & *\textbf{86.54}\% & \textbf{80.39}\% \\
      k-NN (fine-tuned)   & 71.66\%           & 82.36\% & 79.17\%\\
      \bottomrule
    \end{tabular}
  }
    \caption{\label{tab:g2p}Accuracies on prediction of WALS features with language embeddings fine-tuned on Grapheme-to-Phoneme mapping. Asterisks indicate results significantly higher than both other conditions ($p<0.05$).}
\end{table}

\paragraph{Qualitative results}
We now turn to why this task is not a good proxy of phonology.
The task of grapheme-to-phoneme is more related to the processes in the diachronic development of the writing system of a language than it is to the actual genealogy or phonology of the language. 
This is evident when considering the Scandinavian languages Norwegian and Danish which are typologically closely related, and have almost exactly the same orthography.
In spite of this fact, the phonology of the two languages differs drastically due to changes in Danish phonology, which impacts the mapping from graphemes to phonemes severely.
Hence, the written forms of the two languages should be very similar, which makes the language embeddings based on language modelling highly similar to one another.
However, when the embeddings are fine-tuned on a task taking orthography as well as phonology into account, this is no longer the case. 
\begin{figure}[htbp]
	\centering
	\includegraphics[width=0.48\textwidth]{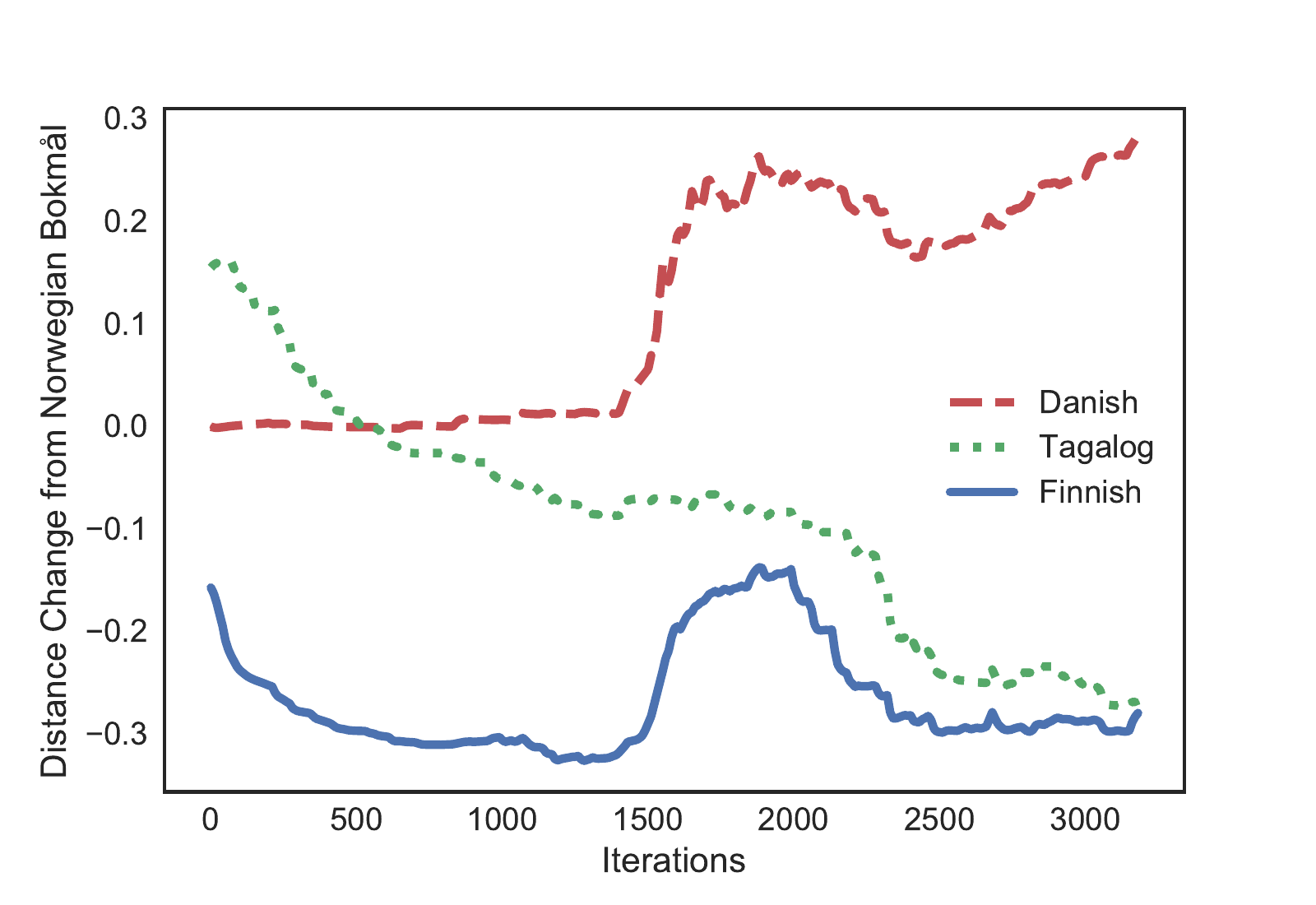}
    \caption{\label{fig:langsim_qual_g2p}Language similarities between Norwegian, and Danish/Tagalog/Finnish, as G2P-based embeddings are fine-tuned.}
\end{figure}
Figure~\ref{fig:langsim_qual_g2p} shows that the language embeddings of Norwegian Bokmål and Danish diverge from each other, which is especially striking when comparing to the converging with the typologically much more distant languages Tagalog and Finnish which.
However, the absolute difference between Norwegian Bokmål and both Tagalog/Finnish is still greater than that of Norwegian Bokmål and Danish even after 3,000 iterations.

\subsection{Phonological reconstruction}
\label{sec:asjp}
As a second phonological task, we look at phonological reconstruction using word lists from the Automated Similarity Judgement Program (ASJP, \citet{asjp}).
This resource contains word lists of at least 40 words per language for more than 4,500 languages.
The task we consider is to reproduce a given source phonological form, also given the language, for instance:
\begin{Verbatim}[fontsize=\small]
(English, wat3r) -> wat3r
\end{Verbatim}
The intuition behind these experiments is that languages with similar phonetic inventories will be grouped together, as reflected in changes in the language embeddings.

\subsubsection{Experiments and Analysis}
We train a sequence-to-sequence model with attention, framed as an auto-encoding problem, using the same sequence-to-sequence architecture and setup as for the grapheme-to-phoneme task.
The model takes as input the characters of each source form together with the language embedding for the language at hand and outputs the predicted target form which is identical to the source form.

\paragraph{Quantitative results}
Since we also consider phonological reconstruction to be a phonological task, we focus the quantitative evaluation on phonological features from WALS.
As with the G2P experiments, Table~\ref{tab:asjp} shows that the fine-tuned embeddings do not offer predictive power above the most frequent class baseline ($p<0.01$).
Observing the changes in the language embeddings reveals that the embeddings are updated to a very small extent, indicating that these are not used by the model to a large extent.
This can be explained by the fact that the task is highly similar for each language, and that the model largely only needs to learn to copy the input string.

We do, however find that evaluating on a set with an unseen language family does yield results significantly above baseline levels with the pre-trained embeddings ($p>0.01$), which together with the G2P results indicate that the language modelling objective does encode features to some extent related to phonology.

\begin{table}[htbp]
  \centering
  \resizebox{\columnwidth}{!}{
    \begin{tabular}{lrrr}
      \toprule
      \textbf{System / features} & \textbf{Random phon.} & \textbf{Unseen phon.} & \textbf{All feat.} \\
      \midrule
      Most Frequent Class & *\textbf{59.39}\% & 63.71\% &  *\textbf{58.12}\% \\
      k-NN (pre-trained)  & 53.02\% & *\textbf{77.44}\% & 51.6\% \\
      k-NN (fine-tuned)   & 53.09\% & *\textbf{77.45}\% & 51.9\%\\
      \bottomrule
    \end{tabular}
  }
    \caption{\label{tab:asjp}Accuracies on prediction of WALS features with language embeddings fine-tuned on Phonological Reconstruction. Asterisks indicate results significantly higher than non-bold conditions ($p<0.01$).}
\end{table}


\section{Morphology}
\subsection{Morphological inflection}
We use data from the Unimorph project, specifically the data released for the SIGMORPHON-2017 shared task \citep{unimorph}.\footnote{\url{https://unimorph.github.io/}}
This data covers 52 languages, thereby representing a relatively large typological variety.
Whereas the shared task has two subtasks, namely inflection and paradigm cell filling, we only train our system using the inflection task. 
This was a choice of convenience, as we are not interested in solving the task of morphological paradigm cell filling, but rather observing the language embeddings as they are fine-tuned.
Furthermore we focus on the high-resource setting in which we have access to 10,000 training examples per language.
The inflection subtask is to generate a target inflected form given a lemma with its part-of-speech as in the following example:
\begin{Verbatim}[fontsize=\small]
(release, V;V.PTCP;PRS) -> releasing
\end{Verbatim}

\subsubsection{Morphological experiments}
\label{sec:sigmorphon}
We train a sequence-to-sequence model with attention over 600 iterations, using the same sequence-to-sequence architecture from the previous tasks.

\paragraph{Quantitative results}
Since this is considered a morphological task, Table~\ref{tab:morph} contains results using the language embeddings to predict morphological properties.
The fine-tuned language embeddings in this condition are able to predict morphological properties in WALS significantly above baseline levels and pre-trained embeddings ($p<0.01$).
We further also obtain significantly better results in the unseen setting ($p<0.01$), in which the language family evaluated on is not used in training.
This indicates that these properties are important to the model when learning the task at hand.

\begin{table}[htbp]
  \centering
  \resizebox{\columnwidth}{!}{
    \begin{tabular}{lrrr}
      \toprule
      \textbf{System / Features} & \textbf{Random morph.} & \textbf{Unseen morph.} & \textbf{All feat.} \\
      \midrule
      Most Frequent Class             & 77.98\%             & 85.68\% & 84.12\% \\
      k-NN (pre-trained)              & 74.49\%             & 88.83\% & \textbf{84.97\%} \\
      k-NN (fine-tuned)               & *\textbf{82.91\%}   & *\textbf{91.92}\% & 84.95\% \\
      \bottomrule
    \end{tabular}
  }
  \caption{\label{tab:morph}Accuracies on prediction of WALS features with language embeddings fine-tuned on morphological inflection.  Asterisks indicate results significantly higher than both other conditions ($p<0.01$).}
\end{table}

\begin{figure}[htbp]
	\centering
	\includegraphics[width=0.243\textwidth,trim={0 0 0.6cm 0},clip]{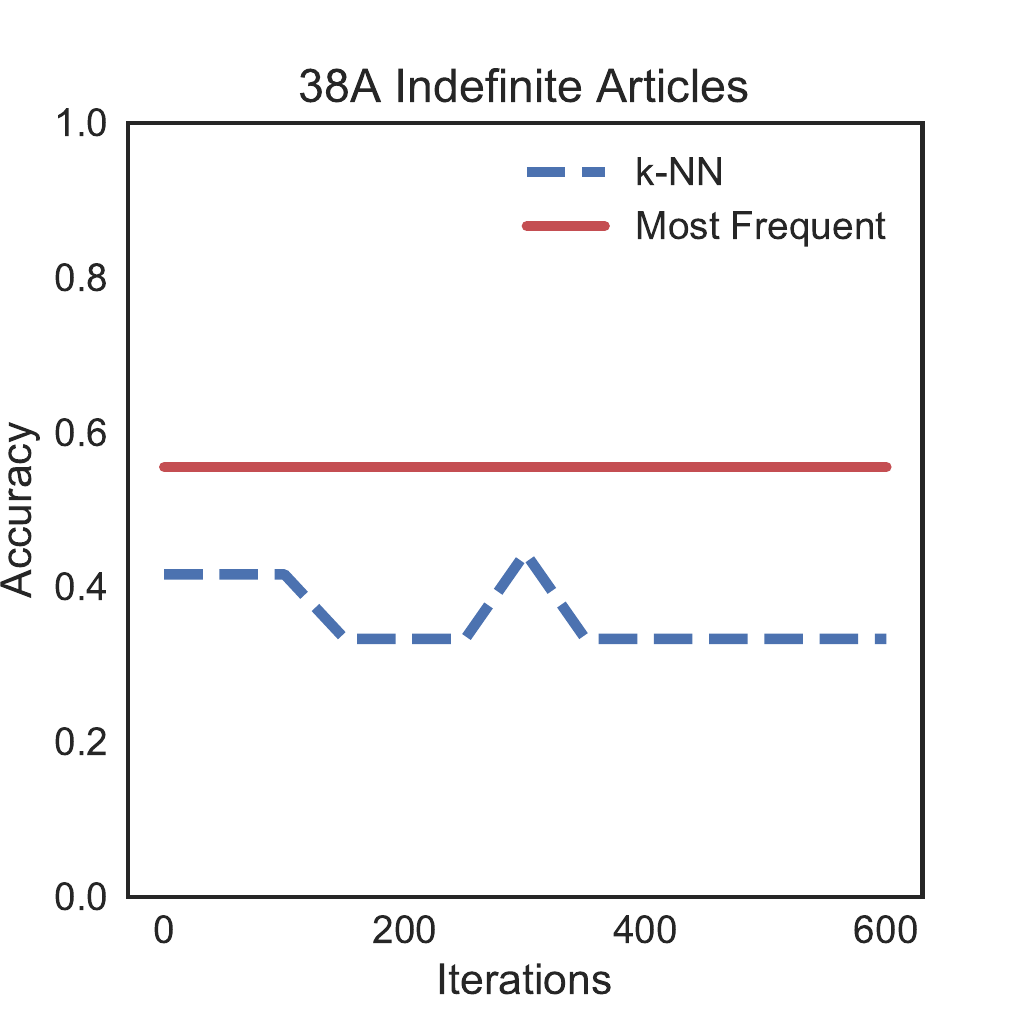}
    \includegraphics[width=0.23\textwidth,trim={0.5cm 0 0.6cm 0},clip]{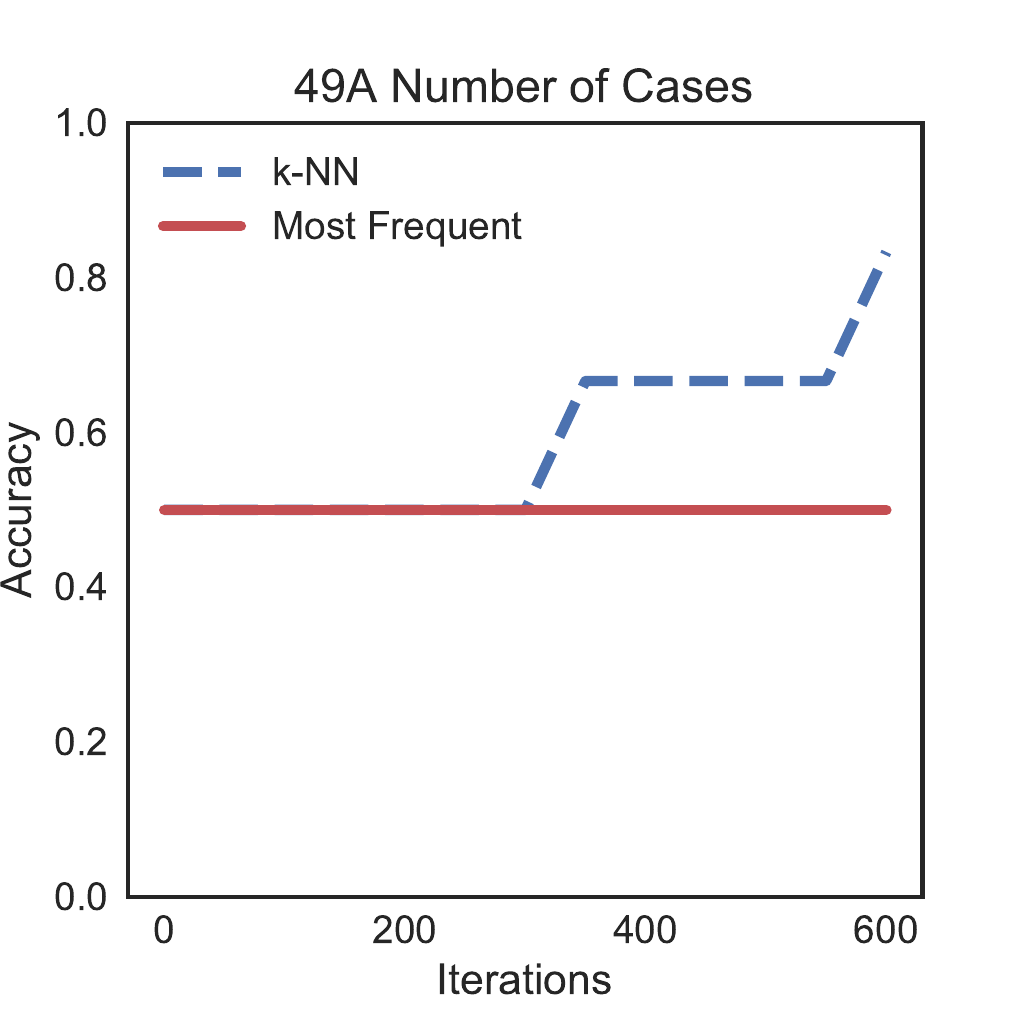}
    \caption{\label{fig:wals_features_morph}Prediction of two morphological features in WALS with morphological language embeddings, one data point per 50 iterations.}
\end{figure}

\paragraph{Qualitative results}
The performance of the fine-tuned embeddings on prediction of morphological features is above baseline for most features.
For $18$ out of the $35$ features under consideration both the baseline and k-NN performances are at $100\%$ from the outset, so these are not considered here.\footnote{This is partially explained by the fact that certain categories were completely uniform in the small sample as well as by the Indo-European bias in the sample.}
Figure~\ref{fig:wals_features_morph} shows two of the remaining $17$ features.\footnote{The full feature set is included in the Supplements.} 
We can observe two main patterns: 
For some features such as 49A (Number of cases), fine-tuning on morphological inflection increases the degree to which the features are encoded in the language embeddings.
This can be explained by the fact that the number of cases in a language is central to how morphological inflection is treated by the model.
For instance, languages with the same number of cases might benefit from sharing certain parameters.
On the other hand, the feature 38A (Indefinite Articles) mainly encodes whether the indefinite word is distinct or not from the word for \textit{one}, and it is therefore not surprising that this is not learned in morphological inflection.

\section{Word order}
\subsection{Part-of-speech tagging}
We use PoS annotations from version 2 of the Universal Dependencies \citep{nivre:2016}.
As we are mainly interested in observing the language embeddings, we down-sample all training sets to 1,500 sentences (approximate number of sentences of the smallest data sets) so as to minimise any size-related effects.

\subsubsection{Word-order experiments}
We approach the task of PoS tagging using a fairly standard bi-directional LSTM architecture based on \citet{plank:2016}, detailed in Section~\ref{sec:implementation}.



\paragraph{Quantitative results}
Table~\ref{tab:pos} contains results on WALS feature prediction using language embeddings fine-tuned on PoS tagging.
We consider both the set of word order features, which are relevant for the dataset, and a set of all WALS features.
Using the fine-tuned embeddings is significantly better than both the baseline and the pre-trained embeddings ($p<0.05$), in both the random and the unseen conditions, indicating that the model learns something about word order typology.
This can be expected, as word order features are highly relevant when assigning a PoS tag to a word.

\begin{table}[htbp]
  \centering
  \resizebox{\columnwidth}{!}{
    \begin{tabular}{lrrr}
      \toprule
      \textbf{System / features} & \textbf{Random W-Order} & \textbf{Unseen W-Order} & \textbf{All feat.} \\
      \midrule
      Most Frequent Class & 76.81\%             & 82.47\% &  82.93\% \\
      k-NN (pre-trained)  & 76.66\%             & 92.76\% &  82.69\% \\
      k-NN (fine-tuned)   & *\textbf{80.81}\%   & *\textbf{94.48}\% &  \textbf{83.55\%} \\
      \bottomrule
    \end{tabular}
  }
  \caption{\label{tab:pos}Accuracies on prediction of WALS features with language embeddings fine-tuned on PoS tagging. Asterisks indicate the result in bold significantly out-performing both other conditions ($p<0.05$).}
\end{table}

\paragraph{Qualitative results}
We now turn to the syntactic similarities between languages as encoded in the fine-tuned language embeddings.
We consider a set of the North-Germanic languages Icelandic, Swedish, Norwegian Nynorsk, Danish, Norwegian Bokmål, the West-Germanic language English, and the Romance languages Spanish, French, and Italian.
We apply hierarchical clustering using UPGMA \citep{upgma} to the pre-trained language embeddings of these languages.\footnote{Included in the Supplements due to space restrictions.}
Striking here is that English is grouped together with the Romance languages.
This can be explained by the fact that English has a large amount of vocabulary stemming from Romance loan words, which under the task of language modelling results in a higher similarity with such languages.
We then cluster the embeddings fine-tuned on PoS tagging in the same way.
In this condition, English has joined the rest of the Germanic languages' cluster.
This can be explained by the fact that, in terms of word ordering and morpho-syntax, English is more similar to these languages than it is to the Romance ones.

We can also observe that, whereas the orthographically highly similar Norwegian Bokmål and Danish form the first sub-cluster in the pre-trained condition, Norwegian Nynorsk replaces Danish in this pairing when fine-tuning on PoS tagging.
This can be explained by the fact that morpho-syntactic similarities between the two written varieties of Norwegian are more similar to one another.

\section{Implementation}
\label{sec:implementation}
\subsection{Sequence-to-sequence modelling}
\begin{figure}[htbp]
	\centering
	\includegraphics[width=0.48\textwidth]{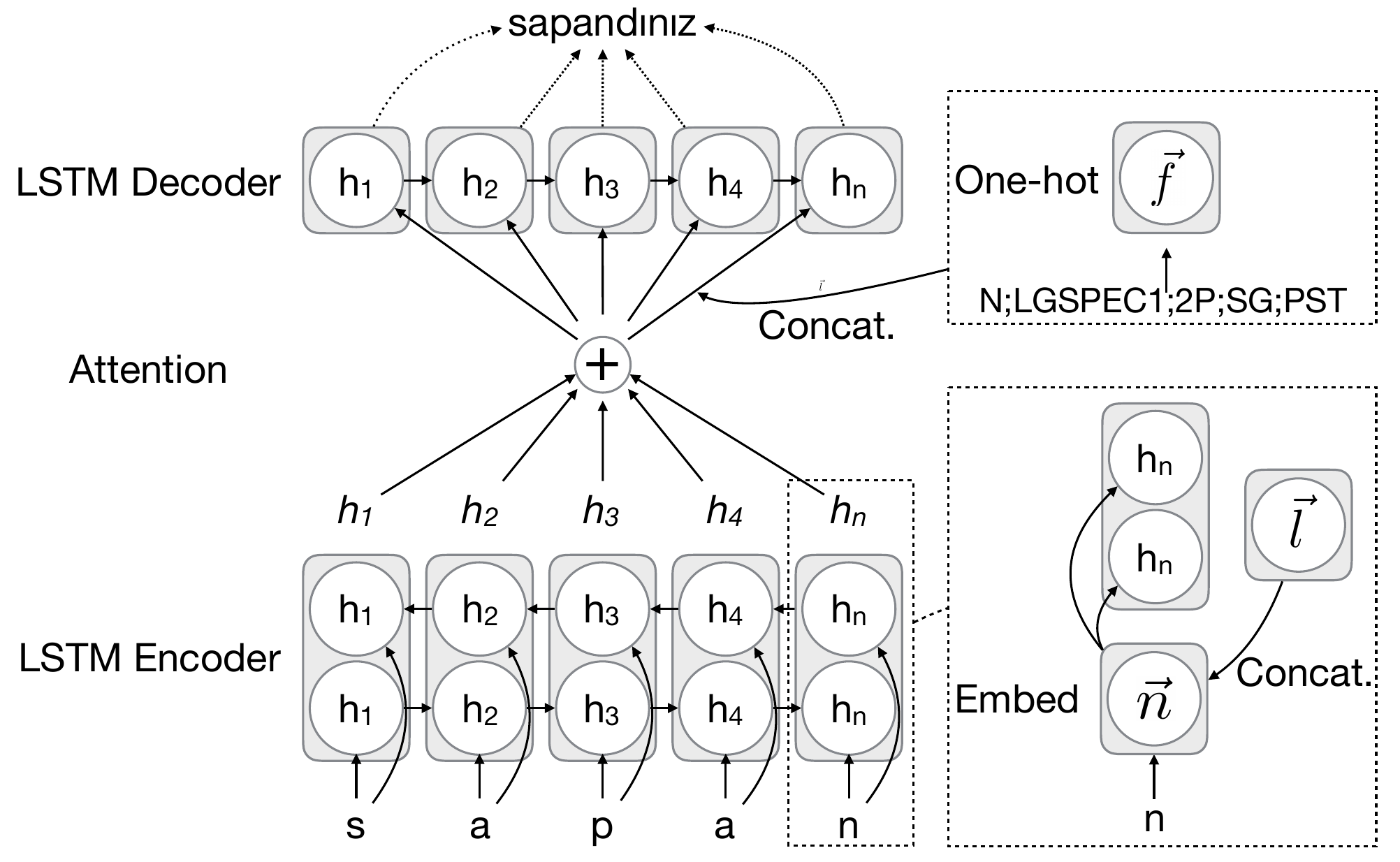}
    \caption{\label{fig:system_architecture}System architecture used in the seq-to-seq tasks (morphological inflection, G2P, and phonological reconstruction).  Figure adapted with permission from \citet{ostling-bjerva:2017}.}
\end{figure}

The system architecture used in the sequence-to-sequence tasks, i.e., G2P, phonological reconstruction, and morphological inflection is depicted in Figure~\ref{fig:system_architecture}.
The system is based on that developed by \citet{ostling-bjerva:2017} and is implemented using Chainer \citep{chainer}. 
We modify the architecture by concatenating a language embedding, $\vec{l}$, to the character embeddings before encoding. 
In the grapheme-to-phoneme and phonological reconstruction experiments, the one-hot feature mapping before decoding is irrelevant and therefore omitted.
The rest of the hyper-parameters are the same as in \citet{ostling-bjerva:2017}.

\subsection{Sequence labelling}
This system is based on \citet{plank:2016}, and is implemented using DyNet \citep{dynet}.
We train using the Adam optimisation algorithm \citep{adam} over a maximum of 10 epochs using early stopping. 
We make two modifications to the bi-LSTM architecture of  \citet{plank:2016}.
First of all, we do not use any atomic embedded word representations, but rather use only character-based word representations.
This choice was made so as to encourage the model not to rely on language-specific vocabulary.
Additionally, we concatenate a pre-trained language embedding to each word representation.
In our formulation, each word $w$ is represented as $LSTM_{c}(w)+\vec{l}$, where $LSTM_{c}(w)$ is the final states of a character bi-LSTM running over the characters in a word and $\vec{l}$ is an embedded language representation.
We use a two-layer deep bi-LSTM with 100 units in each layer, and 100-dimensional character embeddings.
The rest of the hyper-parameters are the same as in \citet{plank:2016}.\footnote{Both modified systems are included in the Supplements, and will be made publicly available.}

\section{Discussion and Conclusions}
The language embeddings obtained by fine-tuning on linguistic tasks at various typological levels were found to include typological information somehow related to the task at hand.
This lends some support to the theoretical foundations of such representations, in that it shows that it is possible to learn something akin to a vector of continuous Chomskyan parameters 
\citep{chomsky:1993}.

\subsection{RQ1: Encoding of typological features in task-specific language embeddings}
The features which are encoded depend to a large degree on the task at hand.
The language embeddings resulting from the phonological tasks did not encode phonological properties in the sense of WALS features, whereas the pre-trained ones did.
The morphological language embeddings were found to encode morphological features, and the PoS language embeddings were similarly found to encode word order features.

A promising result is the fact that we were able to predict typological features for unseen language families.
That is to say, without showing, e.g., a single Austronesian training instance to the k-NN classifier, typological features could still be predicted with high accuracies.
This indicates that we can predict typological features with language embeddings, even for languages for which we have no prior typological knowledge.

Table~\ref{tab:comparison} contains a comparison of the top five and bottom five feature prediction accuracies for the ASJP task.\footnote{See the Supplement for the remaining tasks.} 
In the case of the phonologically oriented ASJP task it is evident that the embeddings still encode something related to phonology, as four out of five best features are phonological.






\subsection{RQ2: Change in encoding of typological features}
The changes in the features encoded in language embeddings are relatively monotonic.
Features are either learnt, forgotten, or remain static throughout training.
This indicates that the language representations converge towards a single optimum.

\subsection{RQ3: Language similarities}
Training language embeddings in the task of multilingual language modelling has been found to reproduce trees which are relatively close matches to more traditional genealogical trees \citep{ostling_tiedemann:2017}.
We show a similar analysis considering pre-trained and PoS fine-tuned embeddings, and it is noteworthy that fine-tuning on PoS tagging in this case yielded a tree more faithful to genealogical trees, such as those represented on \url{glottolog.org}. 

Another striking result in terms of language similarities in fine-tuned language embedding spaces was found in the G2P task.
We here found that the phonological differences between some otherwise similar languages, such as Norwegian Bokmål and Danish, were accurately encoded.



\begin{table}[htbp]
\centering
\small
  
  \begin{tabular}{lrlr}
  \toprule
  \textbf{Task} & \textbf{Feature} & \textbf{WALS Chapter} & \textbf{Accuracy} \\
  \midrule
  \multirow{10}{*}{\bf ASJP}     
 & 6A & Phonology &89.45 \\
 & 18A & Phonology &88.91 \\
 & 20A & Morphology &82.74 \\
 & 19A & Phonology &82.58 \\
 & 7A & Phonology &80.97 \\
 \cmidrule{2-4}
 & 144A & Word Order & 14.48 \\
 & 144L &Word Order & 10.07 \\
 & 62A & Nominal Syntax & 10.00 \\
 & 81B & Word Order & 9.52 \\
 & 133A & Lexicon & 8.18 \\
  \bottomrule
  \end{tabular}
  \caption{\label{tab:comparison}Top 5 and bottom 5 accuracies in feature prediction using phonological language embeddings.}
\end{table}



\bibliographystyle{acl_natbib}
\bibliography{naaclhlt2018}

\end{document}